\documentclass{article}

\usepackage{microtype}
\usepackage{graphicx}
\usepackage{subfigure}
\usepackage{booktabs} 
\usepackage{amsmath}
\usepackage{mathtools}
\usepackage{amssymb}
\usepackage{bm}
\usepackage{multirow}
\usepackage{bbm}
\usepackage{enumitem}
\usepackage[usenames,dvipsnames]{xcolor}
\usepackage{hyperref}

\usepackage{cleveref}
\crefformat{section}{\S#2#1#3} 
\crefformat{subsection}{\S#2#1#3}
\crefformat{subsubsection}{\S#2#1#3}

\usepackage{ragged2e}

\newcommand{\x}{\bm{\mathrm{x}}}
\newcommand{\y}{\bm{\mathrm{y}}}
\newcommand{\z}{\bm{\mathrm{z}}}
\newcommand{\MU}{\bm{\mathrm{\mu}}}

\newcommand{\dec}{p_\theta (\y|\z,\x)}
\newcommand{\pri}{p_\theta (\z|\x)}
\newcommand{\post}{q_\phi (\z|\y,\x)}
\newcommand{\elbobig}{ \E_{\z \sim q_\phi} \Big[ \log \dec \Big] - \text{KL}\Big[ \post \, \Big|\Big| \, \pri \Big]}

\DeclareMathOperator*{\E}{\mathbb{E}}
\newcommand{\wmtende}{WMT'14 En$\rightarrow$De}
\newcommand{\wmtdeen}{WMT'14 De$\rightarrow$En}
\newcommand{\wmtendeboth}{WMT'14 En$\leftrightarrow$De}
\newcommand{\wmtenro}{WMT'16 En$\rightarrow$Ro}
\newcommand{\wmtroen}{WMT'16 Ro$\rightarrow$En}
\newcommand{\wmtenroboth}{WMT'16 En$\leftrightarrow$Ro}
\newcommand{\iwsltdeen}{IWSLT'16 De$\rightarrow$En}
\newcommand{\argmax}{\text{argmax}}
\newcommand{\mygeq}{{}}
\newcommand{\modelts}{{Tr-S}}
\newcommand{\modeltb}{{Tr-B}}
\newcommand{\modeltl}{{Tr-L}}
\newcommand{\modelgb}{{Ga-B}}
\newcommand{\modelgl}{{Ga-L}}
\newcommand{\modelfs}{{Fl-S}}
\newcommand{\modelfb}{{Fl-B}}
\newcommand{\modelfl}{{Fl-L}}

\newcommand{\papertitle}{On the Discrepancy between Density Estimation and Sequence Generation}

\usepackage[preprint]{icml2020}

\icmltitlerunning{\papertitle}

\begin{document}

\twocolumn[
\icmltitle{\papertitle}




\begin{icmlauthorlist}
\icmlauthor{Jason Lee}{nyu}
\icmlauthor{Dustin Tran}{ggl}
\icmlauthor{Orhan Firat}{ggl}
\icmlauthor{Kyunghyun Cho}{nyu,fb,cifar}
\end{icmlauthorlist}

\icmlaffiliation{nyu}{New York University}
\icmlaffiliation{ggl}{Google AI}
\icmlaffiliation{fb}{Facebook AI Research}
\icmlaffiliation{cifar}{CIFAR Associate Fellow}

\icmlcorrespondingauthor{Jason Lee}{jason@cs.nyu.edu}

\icmlkeywords{vae, flow, normalizing flow, prior, machine translation, seq2seq, sequence generation, sequence modeling}

\vskip 0.3in
]



\printAffiliationsAndNotice{} 

\begin{abstract}
Many sequence-to-sequence generation tasks, including machine translation and text-to-speech, can be posed as estimating the density of the output $y$ given the input $x$: $p(y|x).$ Given this interpretation, it is natural to evaluate sequence-to-sequence models using conditional log-likelihood on a test set. 
However, the goal of sequence-to-sequence generation (or structured prediction) is to find the best output $\hat{y}$ given an input $x$, and each task has its own downstream metric $R$ that scores a model output by comparing against a set of references $y^*$: $R(\hat{y}, y^* | x).$
While we hope that a model that excels in density estimation also performs well on the downstream metric, the exact correlation has not been studied for sequence generation tasks.
In this paper, by comparing several density estimators on five machine translation tasks, we find that the correlation between rankings of models based on log-likelihood and BLEU varies significantly depending on the range of the model families being compared.
First, log-likelihood is highly correlated with BLEU when we consider models within the same family (e.g. autoregressive models, or latent variable models with the same parameterization of the prior).
However, we observe no correlation between rankings of models across different families:
(1) among non-autoregressive latent variable models, a flexible prior distribution is better at density estimation but gives worse generation quality than a simple prior, and (2) autoregressive models offer the best translation performance overall, while latent variable models with a normalizing flow prior give the highest held-out log-likelihood across all datasets.
Therefore, we recommend using a simple prior for the latent variable non-autoregressive model when fast generation speed is desired.
\end{abstract}

\section{Introduction}
\label{intro}

Sequence-to-sequence generation tasks can be cast as conditional density estimation $p(y|x)$ where $x$ and $y$ are input and output sequences. In this framework, density estimators are trained to maximize the conditional log-likelihood, and also evaluated using log-likelihood on a test set. 
However, many sequence generation tasks require finding the best output $\hat{y}$ given an input $x$ at test time, and the output is evaluated against a set of references $y^*$ on a task-specific metric: $R(\hat{y},y^*|x).$
For example, machine translation systems are evaluated using BLEU scores~\citep{papieni02bleu}, image captioning systems use METEOR~\citep{banerjee05meteor} and text-to-speech systems use MOS (mean opinion scores). 
As density estimators are optimized on log-likelihood, we want models with higher held-out log-likelihoods to give better generation quality, but the correlation has not been well studied for sequence generation tasks. 
In this work, we investigate the correlation between rankings of density estimators based on (1) test log-likelihood and (2) the downstream metric for machine translation~\footnote{We open source our code at \url{https://github.com/tensorflow/tensor2tensor}}.

On five language pairs from three machine translation datasets ({\wmtendeboth}, {\wmtenroboth}, {\iwsltdeen}), we compare the held-out log-likelihood and BLEU scores of several density estimators: (1) autoregressive models~\citep{vaswani17attention}, (2) latent variable models with a non-autoregressive decoder and a simple (diagonal Gaussian) prior~\citep{shu19latent}, and (3) latent variable models with a non-autoregressive decoder and a flexible (normalizing flow) prior~\citep{ma19flowseq}.

We present two key observations. First, among models within the same family, we find that log-likelihood is strongly correlated with BLEU. The correlation is almost perfect for autoregressive models and high for latent variable models with the same prior. Between models of different families, however, log-likelihood and BLEU are not correlated. Latent variable models with a flow prior are in fact the best density estimators (even better than autoregressive models), but they give the worst generation quality. Gaussian prior models offer comparable or better BLEU scores, while autoregressive models give the best BLEU scores overall.
From these findings, we conclude that the correlation between log-likelihood and BLEU scores varies significantly depending on the range of model families considered.

Second, we find that knowledge distillation drastically hurts density estimation performance across different models and datasets, but consistently improves translation quality of non-autoregressive models. 
For autoregressive models, distillation slightly hurts translation quality. Among latent-variable models, iterative inference with a delta posterior~\citep{shu19latent} significantly improves the translation quality of latent variable models with a Gaussian prior, whereas the improvement is relatively small for the flow prior.
Overall, for fast generation, we recommend a latent variable non-autoregressive model using a simple prior (rather than a flexible one), knowledge distillation, and iterative inference. This is 5--7x faster than the autoregressive model at the expense of 2 BLEU scores on average, and it improves upon latent variable models with a flexible prior across generation speed, BLEU, and parameter count.

\section{Background}
\label{sec:background}
Sequence-to-sequence generation is a supervised learning problem of generating an output sequence given an input sequence. For many such tasks, conditional density estimators have been very successful~\citep{sutskever14sequence,bahdanau15neural,vinyals15show,vinyals15neural}. 

To learn the distribution of an output sequence, it is crucial to give enough capacity to the model to be able to capture the dependencies among the output variables. We explore two ways to achieve this: (1) directly modeling the dependencies with an autoregressive factorization of the variables, and (2) letting latent variables capture the dependencies, so the distribution of the output sequence can be factorized given the latent variables and therefore more quickly be generated.
We discuss both classes of density estimators in depth below. We denote the training set as a set of tuples $\{(\x_n, \y_n)\}_{n=1}^N$ and each input and output example as sequences of random variables $\x=\{x_1, \dots, x_{T'}\}$ and $\y=\{y_1, \dots, y_{T}\}$ (where we drop the subscript $n$ to reduce clutter). 
We use $\theta$ to denote the model parameters.

\subsection{Autoregressive Models}
\paragraph{Learning}
Autoregressive models factorize the joint distribution of the sequence of output variables $\y=\{y_1,\dots,y_T\}$ as a product of conditional distributions:
$$\log p_{\text{AR}}(\y|\x) = \sum_{t=1}^{T} \log p_\theta(y_t|y_{<t},\x).$$
They are trained to maximize the log-likelihood of the training data: $L_\text{AR}(\theta) = \frac{1}{N} \sum_{n=1}^{N} \log p_{\text{AR}}(\y_n|\x_n).$

\paragraph{Parameterization}
Recurrent neural networks and their gated variants are natural parameterizations of autoregressive models~\citep{elman90finding,hochreiter97long,chung14empirical}. 
By ensuring that no future information $y_{\geq t}$ is used in predicting the current timestep $y_t$, non-recurrent architectures can also parameterize autoregressive models, such as convolutions~\citep{oord16wavenet,gehring17convolutional} and Transformers~\citep{vaswani17attention}, which are feedforward networks with self-attention.

\paragraph{Inference} Finding the most likely output sequence given an input sequence under an autoregressive model amounts to solving a search problem:
$$\argmax_{\y} \log p_\theta (\y|\x) = \argmax_{y_{1:T}} \sum_{t=1}^{T}{\log p_\theta (y_t|y_{<t}, \x)}.$$
As the size of the search space grows exponentially with the length of the output sequence $T$, solving this exactly is intractable. Therefore, approximate search algorithms are often used such as greedy search or beam search.

\subsection{Latent Variable Models}
\paragraph{Learning}
Latent variable models posit a joint distribution of observed variables ($\y$) and unobserved variables ($\z$). They are trained to maximize the marginal log-likelihood of the training data:
\begin{equation}
\log p_{\text{LVM}} (\y|\x) = \log \int_{\z} \dec \: \pri d\z.
\end{equation}
As the marginalization over $\z$ makes computing the marginal log-likelihood and posterior inference intractable, variational inference proposes to use a parameterized family of distributions $q_\phi(\z|\y,\x)$ to approximate the true posterior $p(\z|\y,\x).$ Then, we have the evidence lowerbound (ELBO)~\citep{wainwright08graphical,kingma14auto}:
\begin{align}
\label{eq:elbo}
\log & \: p_{\text{LVM}} (\y|\x) \geq \text{ELBO}(\y,\x;\theta,\phi) \\
    &= \elbobig , \notag
\end{align}
where $\dec$ is the decoder, $\post$ is the variational posterior and $\pri$ is the prior. Both the model and variational parameters $\theta, \phi$ are estimated to maximize ELBO over the training set: $L_{\text{LVM}}(\theta,\phi) = \frac{1}{N}\sum_{n=1}^{N} \text{ELBO}(\y_n,\x_n;\theta,\phi).$

\paragraph{Parameterization}

As latent variables can capture the dependencies between the output variables, the decoding distribution can be factorized: $\dec = \prod_{t=1}^{T} p_\theta(y_t|\z,\x)$. 
The approximate posterior distribution is also often factorized, which can be parameterized by any neural network that outputs mean and standard deviation for each output position:
$q_{\phi}(z_{1:T}|\y,\x) = \prod_{t=1}^{T} \mathcal{N}\Big(z_{t} \Big|\mu_{\phi,t}(\y,\x), \sigma_{\phi,t}(\y, \x)\Big).$
We discuss prior distributions in \cref{sec:lvmprior}.

\paragraph{Inference}
Generating the most likely output given an input with a latent variable model requires optimizing ELBO with respect to the output: $\argmax_{\y}{\text{ELBO}(\y,\x;\theta,\phi)}.$
As computing the expectation in Eq.~\ref{eq:elbo} is intractable, we instead optimize a proxy lowerbound using a delta posterior~\citep{shu19latent}: 
\begin{equation*}
  \delta(\z|\MU) =
    \begin{cases}
      1, & \text{if} \:\:\z = \MU\\
      0, & \text{otherwise}
    \end{cases}       
\end{equation*}
Then, the ELBO reduces to:
\vskip -0.3in
\begin{align}
& \E_{\z\sim \delta(\z|\MU)}{\Big[ \dec + \pri \Big]} + \overbrace{\mathcal{H}(\delta)}^{=0}, \notag \\
&= \log p_{\theta} (\y|\MU, \x) + \log p_{\theta} (\MU|\x).
\label{eq:proxy}
\end{align}

We maximize Eq.~\ref{eq:proxy} with iterative refinement: the EM algorithm alternates between (1) matching the proxy to the original lowerbound by setting ${\MU} = \E_{q_\phi}[\z]$, and (2) maximizing the proxy lowerbound with respect to $\y$ by: $\hat{\y} = \text{argmax}_{\y} (\log p_\theta(\y|{\MU},\x)).$ 
The delta posterior is initialized using the prior (e.g. $\MU=\E_{\z \sim \pri}[\z]$ in case of a Gaussian prior) so that the inference algorithm is fully deterministic, a desirable property for sequence generation tasks. We study the effect of iterative refinement on BLEU score in detail.

\subsection{Prior for Latent Variable Models}
\label{sec:lvmprior}
Several work have discovered that the prior distribution plays a critical role in balancing the variational posterior and the decoder, and a standard normal distribution may be too rigid for the aggregate posterior to match~\citep{hoffman16elbo,rosca18distribution}. Indeed, follow-up work found that more flexible prior distributions outperform simple priors on several density estimation tasks~\citep{tomczak18vae,bauer19resampled}.
Therefore, we explore two choices for the prior distribution: a factorized Gaussian and a normalizing flow.

\paragraph{Diagonal Gaussian}
A simple model of the conditional prior is a factorized Gaussian distribution:
$$\log p_{\theta}(z_{1:T}|\x) = \sum_{t=1}^{T} {\log \mathcal{N}\Big(z_{t} \Big|\mu_{\theta,t}(\x), \sigma_{\theta,t}(\x)\Big),}$$
where each latent variable $z_t$ is modeled as a diagonal Gaussian with mean and standard deviation computed from a learned function.

\paragraph{Normalizing Flow}
Normalizing flows~\citep{tabak13family,rezende15variational,papa19normalizing} offer a general method to construct complex probability distributions over continuous random variables. It consists of (1) a base distribution $p_{b}(\epsilon)$ (often chosen as a standard Gaussian distribution) and an invertible transformation $f$ and its inverse $f^{-1}$, such that $f(\z)=\epsilon,\:\:f^{-1}(\epsilon)=\z.$
As our prior is conditioned on $\x$, so are the transformations: $f(\z; \x)=\epsilon,\:\:f^{-1}(\epsilon; \x)=\z.$
Then, by change-of-variables, we can evaluate the exact density of the latent variable ${\z}$ under the flow prior:
$$\log p_\theta(\z|\x) = \log p_b\Big(f(\z; \x)\Big) + \log \bigg|\text{det} \frac{\partial f(\z; \x)}{\partial \z}\bigg|.$$
Affine coupling flows~\citep{dinh17density} enable efficient generation and computation of the Jacobian determinant by constructing each transformation such that only a subset of the random variables undergoes affine transformation, using parameters computed from the remaining variables:
\begin{align}
\label{eq:glow}
    &\z_\text{id}, \z_\text{tr} = \text{split}(\z) \nonumber \\
    &\bm{\mathrm{s}}, \bm{\mathrm{b}} = g_\text{param}(\z_\text{id}) \\
    &f(\z) = \text{concat}(\z_\text{id}; \:\: \bm{\mathrm{s}} \cdot \z_\text{tr} + \bm{\mathrm{b}}), \nonumber
\end{align}
where $g_\text{param}$ can be arbitrarily complex as it needs not be invertible. 
As invertibility is closed under function composition and the Jacobian determinant is multiplicative, increasingly flexible coupling flows can be constructed by stacking multiple flow layers and reordering such that all the variables are transformed. 

\subsection{Flow-based Density Estimators}
As normalizing flows apply continuous transformations to continuous distributions, they are not directly applicable to discrete data such as text. 
Recently proposed discrete normalizing flows (without the determinant Jacobian term) give promising performance on character-level language modeling and image compression~\citep{tran19discrete,hoogeboom19integer}. However, bias from straight-through gradient estimators hinders scalability in terms of flow depth and the number of classes. As this reduces their chance of success in large-scale sequence generation tasks such as machine translation, we do not include discrete flow models in our experiments and leave it as future work.

\subsection{Knowledge Distillation}
While most density estimators for sequence generation tasks are trained to maximize the log-likelihood of the training data, recent work have shown that it is possible to improve the performance of non-autoregressive models significantly by training them on the predictions of a pre-trained autoregressive model~\citep{gu18non,oord18parallel}. 
While \citet{zhou19understanding} recently found that distillation reduces complexity of the training data, its effect on density estimation performance has not been studied.

\section{Problem Definition}
\label{sec:probdef}
On a sequence generation task, a conditional density estimator $F \in \mathcal{H}$ (where $\mathcal{H}$ is a hypothesis set of density estimators in \cref{sec:background}) is trained to maximize the log-likelihood (or its approximation) of the training set $\{(x_n, y_n)\}_{n=1}^N$:
$$L(F) = \frac{1}{N} \sum_{n=1}^{N} \log p_{F}(y_n|x_n).$$

Once training converges, the model $F$ is evaluated on the test set $\{(x_m, y_m)\}_{m=1}^M$ using a downstream metric $R$:
$$R(F) = R((y_1, \dots, y_M), (\hat{y}_1, \dots, \hat{y}_M), (x_1, \dots, x_M)),$$
where $\hat{y}_m = \argmax_{y} \log p_{F} (y|x_m).$

To perform model selection, we can rank a set of density estimators $\{F_1, \dots, F_K\}$ based on either the held-out log-likelihood or the downstream metric. We measure the correlation between the rankings given by the log-likelihood $L(F)$ and the downstream metric $R(F)$.

\section{Experimental Setup}
On machine translation, we train several autoregressive models and latent variable models and analyze the correlation between their rankings based on log-likelihood and BLEU.

\subsection{Datasets and Preprocessing}
We use five language pairs from three translation datasets: {\iwsltdeen}\footnote{\url{https://wit3.fbk.eu/}} (containing 197K training, 2K development and 2K test sentence pairs), {\wmtenroboth}\footnote{\url{www.statmt.org/wmt16/translation-task.html}} (612K, 2K, 2K pairs) and {\wmtendeboth}\footnote{\url{www.statmt.org/wmt14/translation-task.html}} (4.5M, 3K, 3K pairs). For {\wmtendeboth} and {\wmtenroboth}, both directions are used.

We use the preprocessing scripts with default hyperparameters from the \texttt{tensor2tensor} framework.\footnote{\url{https://github.com/tensorflow/tensor2tensor/blob/master/tensor2tensor/bin/t2t-datagen}} Namely, we use wordpiece tokenization~\citep{schuster12japanese} with 32K wordpieces on all datasets. For {\wmtenroboth}, we follow \citet{sennrich16edinburgh} and normalize Romanian and remove diacritics before applying wordpiece tokenization. For training, we discard sentence pairs if either the source or the target length exceeds 64 tokens. 
As splitting along the time dimension~\citep{ma19flowseq} in the coupling flow layer requires that the length of the output sequence is a multiple of 2 at each level, \texttt{<EOS>} tokens are appended to the target sentence until its length is a multiple of 4.

\subsection{Autoregressive Models}
We use three Transformer~\citep{vaswani17attention} models of different sizes: Transformer-big (\modeltl), Transformer-base (\modeltb) and Transformer-small (\modelts). The first two models have the same hyperparameters as in \citet{vaswani17attention}. Transformer-small has 2 attention heads, 5 encoder and decoder layers, $d_{\text{model}}=256$ and $d_\text{filter}=1024$.

\subsection{Latent Variable Models}
The latent variable models in our experiments are composed of the source sentence encoder, length predictor, prior, decoder and posterior. The source sentence encoder is implemented with a standard Transformer encoder. 
Given the hidden states of the source sentence, the length predictor (a 2-layer MLP) predicts the length difference between the source and target sentences as a categorical distribution in $[-30, 30].$
We implement the decoder $\dec$ with a standard Transformer decoder that outputs the logits of all target tokens in parallel. The approximate posterior $\post$ is implemented as a Transformer decoder with a final Linear layer with weight normalization~\citep{salimans16weight} to output the mean and standard deviation (having dimensionality $d_\text{latent}$). Both the decoder and the approximate posterior attend to the source hidden states.

\paragraph{Diagonal Gaussian Prior}
The diagonal Gaussian prior is implemented with a Transformer decoder which receives a sequence of positional encodings of length $T$ as input, and outputs the mean and standard deviation of each target token (of dimensionality $d_\text{latent}$). We train two models of different sizes: Gauss-base (\modelgb) and Gauss-large (\modelgl). Gauss-base has 4 attention heads, 3 posterior layers, 3 decoder layers and 6 encoder layers, whereas Gauss-large has 8 attention heads, 4 posterior layers, 6 decoder layers, 6 encoder layers.
$(d_\text{model}, d_\text{latent}, d_\text{filter})$ is (512, 512, 2048) for WMT experiments and (256, 256, 1024) for IWSLT experiments.

\paragraph{Normalizing Flow Prior}
The flow prior is implemented with Glow~\citep{kingma18glow}. We use a single Transformer decoder layer with a final Linear layer with weight normalization to parameterize $g_\text{param}$ in Eq.~\ref{eq:glow}. This produces the shift and scale parameters for the affine transformation.
Our flow prior has the multi-scale architecture with three levels~\citep{dinh17density}: at the end of each level, half of the latent variables are modeled with a standard Gaussian distribution.
We use three split patterns and multi-headed 1x1 convolution from \citet{ma19flowseq}. We experiment with the following hyperparameter settings: Flow-small (\modelfs) with 12/12/8 flow layers in each level and Flow-base (\modelfb) with 12/24/16 flow layers in each level.
The first level corresponds to the latent distribution and the last level corresponds to the base distribution. $(d_\text{model}, d_\text{latent}, d_\text{filter})$ is (320, 320, 640) for all experiments. For the Transformer decoder in $g_\text{param}$, we use 4 attention heads for Flow-small and 8 attention heads for Flow-base.


\subsection{Training and Optimization}
We use the Adam~\citep{kingma15adam} optimizer with the learning rate schedule used in \citet{vaswani17attention}. The norm of the gradients is clipped at 1.0. 
We perform early stopping and choose the learning rate warmup steps and dropout rate based on the BLEU score on the development set.
To train non-autoregressive models, the loss from the length predictor is minimized jointly with negative ELBO loss.

\paragraph{Knowledge Distillation} Following previous work~\citep{kim16sequence,gu18non,lee18deterministic}, we construct a distilled dataset by decoding the training set using Transformer-base with beam width 4. For {\iwsltdeen}, we use Transformer-small.

\paragraph{Latent Variable Models} To ease optimization of latent variable models~\citep{bowman16generating,higgins17beta}, we set the weight of the KL term to 0 for the first 5,000 SGD steps and linearly increase it to 1 over the next 20,000 steps. Similarly with \citet{mansimov19molecular}, we find it helpful to add a small regularization term to the training objective that matches the approximate posterior with a standard Gaussian distribution: $\alpha \cdot \text{KL}\big[ q_\phi (\z|\y,\x) \: || \:\mathcal{N}(0, \bm{\mathrm{I}}) \big]$, as the original KL term $\text{KL}\big[ \post \, \big|\big| \, \pri \big]$ does not have a local point minimum but a valley of minima.
We find $\alpha=10^{-4}$ to work best.

\paragraph{Flow Prior Models} We perform data-dependent initialization of actnorm parameters for the flow prior~\citep{kingma18glow} at the 5,000-th step, which is at the beginning of KL scheduling.

\subsection{Evaluation Metrics}

\paragraph{Log-likelihood} is the main metric for measuring density estimation (data modeling) performance. We compute exact log-likelihood for autoregressive models. 
For latent variable models, we estimate the marginal log-likelihood by importance sampling with 1K samples from the approximate posterior and using the ground truth target length.

\paragraph{BLEU} measures the similarity (in terms of n-gram overlap) between a generated output and a set of references, regardless of the model. It is a standard metric for generation quality of machine translation systems. We also compute \textbf{Pairwise BLEU} to measure the diversity among a set of outputs generated from a model~\citep{shen19mixture}.

\paragraph{Generation Speed} In addition to the quality-driven metrics, we measure the generation speed of each model in the number of sentences generated per second on a single V100 GPU with batch size 1.

\section{Results}
\label{sec:results}

\begin{table}[!t]
\small
\centering
\vskip -0.10in
\caption{Test BLEU score and log-likelihood of each model. Raw: models trained on raw data. Dist.: models trained on distilled data. \modelts: Transformer-small. \modeltb: Transformer-base. \modeltl: Transformer-big. \modelgb: Gauss-base. \modelgl: Gauss-large. \modelfs: Flow-small. \modelfb: Flow-base. \modelfl: Flow-large.
We use beam search with width 4 for inference with autoregressive models, and one step of iterative inference~\citep{shu19latent} for latent variable models.
On most datasets, our Flow-base model gives comparable results to those from \citet{ma19flowseq}, which are denoted with ($*$).
We boldface the best log-likelihood overall and the best BLEU score among the latent variable models. We underscore best BLEU score among the autoregressive models.
}
\vskip 0.15in
\begin{sc}
\begin{tabular}{llrrrr} \toprule
 & & \multicolumn{2}{c}{BLEU ($\uparrow$)} & \multicolumn{2}{c}{LL ($\uparrow$)} \\ 
 & & \multicolumn{1}{c}{Raw} & \multicolumn{1}{c}{Dist.} & \multicolumn{1}{c}{Raw} & \multicolumn{1}{c}{Dist.} \\ 
 
\midrule
\multirow{11}{*}{\rotatebox{90}{$\:\:\:$\wmtende}} 
& \modelts & 24.54 & 24.94 & -1.77 & -2.36 \\
& \modeltb & 28.18 & 27.86 & -1.44 & -2.19 \\
& \modeltl & \underline{29.39} & {28.29} & -1.35 & -2.23 \\
\cmidrule{2-6}
& \modelgb & 15.74 & 24.54 &  \mygeq -1.51 & \mygeq -2.44 \\
& \modelgl & 17.33 & \textbf{25.53} &  \mygeq -1.47 & \mygeq -2.24 \\
& \modelfs & 18.17 & 21.98 &  \mygeq -1.41 & \mygeq -2.13 \\
& \modelfb & 18.57 & 21.82 &  \mygeq \textbf{-1.23} & \mygeq -2.05 \\
\cmidrule{2-6}
& \modelfb$^{(*)}$ & 18.55  & 21.45 & & \\
& \modelfl$^{(*)}$ & 20.85  & 23.72 & & \\ 

\midrule
\multirow{11}{*}{\rotatebox{90}{$\:\:\:$\wmtdeen}} 
& \modelts & 29.15 & 28.40 & -1.66 & -2.24 \\
& \modeltb & 32.21 & {32.24} & -1.42 & -2.12 \\
& \modeltl & \underline{33.16} & {32.24} & -1.35 & -2.05 \\
\cmidrule{2-6}
& \modelgb & 21.64 & 29.29 &  \mygeq -1.41 & \mygeq -2.17 \\
& \modelgl & 23.03 & \textbf{30.30} &  \mygeq -1.31 & \mygeq -2.04  \\
& \modelfs & 23.17 & 27.14 &  \mygeq -1.28 & \mygeq -1.73  \\
& \modelfb & 23.12 & 26.72 &  \mygeq \textbf{-1.20} & \mygeq -1.71  \\
\cmidrule{2-6}
& \modelfb$^{(*)}$ & 23.36  & 26.16 & &   \\
& \modelfl$^{(*)}$ & 25.40  & 28.39 & &   \\ 

\midrule
\multirow{10}{*}{\rotatebox{90}{$\:\:\:$\wmtenro}} 
& \modelts & 30.12 & 29.57 & -1.72 & -1.95 \\
& \modeltb & \underline{33.46} & {33.28} & -1.63 & -2.52  \\
\cmidrule{2-6}
& \modelgb & 28.03 & 29.71 &  \mygeq -2.38 & \mygeq -3.48  \\
& \modelgl & 28.16 & \textbf{30.91} &  \mygeq -2.44 & \mygeq -3.54  \\
& \modelfs & 26.85 & 28.63 &  \mygeq -1.53 & \mygeq -2.42  \\
& \modelfb & 27.49 & 29.09 &  \mygeq \textbf{-1.50} & \mygeq -2.31  \\
\cmidrule{2-6}
& \modelfb$^{(*)}$ & 29.26 & 29.34 & &   \\
& \modelfl$^{(*)}$ & 29.86 & 29.73 & &   \\ 

\midrule
\multirow{10}{*}{\rotatebox{90}{$\:\:\:$\wmtroen}} 
& \modelts & 29.33 & 28.87 & -1.84 & -1.93 \\
& \modeltb & \underline{32.19} & {31.15} & -1.79 & -2.28  \\
\cmidrule{2-6}
& \modelgb & 26.48 & 27.81 &  \mygeq -2.41 & \mygeq -2.92  \\
& \modelgl & 27.35 & \textbf{28.02} &  \mygeq -2.32 & \mygeq -3.01  \\
& \modelfs & 26.03 & 26.12 &  \mygeq -1.65 & \mygeq -2.05  \\
& \modelfb & 27.14 & 27.33 &  \mygeq \textbf{-1.64} & \mygeq -2.01  \\
\cmidrule{2-6}
& \modelfb$^{(*)}$ & 30.16 & 30.44 & &   \\
& \modelfl$^{(*)}$ & 30.69 & 30.72 & &   \\ 

\midrule
\multirow{9}{*}{\rotatebox{90}{$\quad\quad\quad$IWSLT}} 
& \modelts & 31.54 & \underline{31.72} & -1.84 & -2.56  \\
\cmidrule{2-6}
& \modelgb & 24.36 & 26.80 &  \mygeq -1.98 & \mygeq -2.70  \\
& \modelfs & 23.64 & 26.69 &  \mygeq -1.66 & \mygeq -2.28  \\
& \modelfb & 24.89 & \textbf{27.00} &  \mygeq \textbf{-1.57} & \mygeq -2.46  \\
\cmidrule{2-6}
& \modelfb$^{(*)}$ & 24.75 & 27.75 & &   \\

\bottomrule
\end{tabular}
\end{sc}
\vskip -0.40in
\label{tab:mainresult}
\end{table}

\begin{table}[!t]
\small
\centering
\vskip -0.1in
\caption{Pearson's correlation between log-likelihood and BLEU across the training checkpoints of Transformer-base, Gauss-base and Flow-base on {\wmtende}.}
\vskip 0.15in
\begin{sc}
\begin{tabular}{lrrr} \toprule
 & \modeltb & \modelgb & \modelfb \\ \midrule
Raw & 0.926 & 0.831 & 0.678 \\
Dist. & -0.758 & -0.897 & -0.873 \\
\bottomrule
\end{tabular}
\end{sc}
\label{tab:corr}
\end{table}

\begin{table}[!t]
\small
\centering
\vskip -0.1in
\caption{BLEU scores and log-likelihoods on out-of-distribution test sets. Models trained on {\wmtdeen} are evaluated on {\iwsltdeen}, and vice versa.}
\vskip 0.15in
\begin{sc}
\begin{tabular}{llrrrr} \toprule
 & & \multicolumn{2}{c}{BLEU ($\uparrow$)} & \multicolumn{2}{c}{LL ($\uparrow$)} \\ 
 & & \multicolumn{1}{c}{Raw} & \multicolumn{1}{c}{Dist.} & \multicolumn{1}{c}{Raw} & \multicolumn{1}{c}{Dist.} \\ 
 
\midrule
\multirow{8}{*}{\rotatebox{90}{\parbox{1.2cm}{\centering WMT'14 \\ $\rightarrow$IWSLT}}} 
    & \modelts & 29.15 & 28.40 & -1.65 & -2.25 \\
    & \modeltb & 32.29 & 31.75 & -1.42 & -2.12 \\
    & \modeltl & \underline{33.16} & 32.24 & -1.35 & -2.06 \\
\cmidrule{2-6}
& \modelgb & 24.26 & 28.77 & \mygeq -1.37 & \mygeq -2.10 \\
& \modelgl & 25.46 & \textbf{29.60} & \mygeq -1.28 & \mygeq -2.01 \\
& \modelfs & 24.35 & 26.79 & \mygeq -1.26 & \mygeq -1.76 \\
& \modelfb & 24.25 & 27.12 & \mygeq \textbf{-1.19} & \mygeq -1.73 \\

\midrule
\multirow{6}{*}{\rotatebox{90}{\parbox{0.8cm}{\centering IWSLT$\rightarrow$ \\ WMT'14}}} 
    & \modelts & 18.50 & \underline{18.94} & -2.79 & -3.41 \\
\cmidrule{2-6}
& \modelgb & 12.12 & 13.78 & \mygeq -3.10 & \mygeq -3.83 \\
& \modelfs & 11.78 & \textbf{14.35} & \mygeq -2.81 & \mygeq -3.22 \\
& \modelfb & 12.56 & 14.30 & \mygeq \textbf{-2.62} & \mygeq -3.43 \\

\bottomrule
\end{tabular}
\end{sc}
\label{tab:oodresult}
\end{table}

\subsection{Correlation between rankings of models}
\label{sec:mainresult}

Table~\ref{tab:mainresult} presents the comparison of three model families (Transformer, Gauss, Flow) on five language pairs in terms of generation quality (BLEU) and log-likelihood (LL). We present two sets of results: one from models trained on raw data (Raw), and another from models trained on distilled data (Dist.) (which we mostly discuss in \cref{sec:dist}). 
We use the original test set in computing the log-likelihood and BLEU scores of the distilled models, so the results are comparable with the undistilled models.
We make two main observations:

\begin{enumerate}[leftmargin=*]
    \vskip -0.1in
    \setlength\itemsep{0.40em}
    \setlength\parskip{0.0em}
    \item Log-likelihood is highly correlated with BLEU when considering models within the same family.
    \begin{enumerate}[
      align=left,
      leftmargin=1.0em,
      itemindent=0pt,
      labelsep=0pt,
      labelwidth=2em
    ]
    \setlength\itemsep{0.40em}
    \setlength\parskip{0.0em}
        \item Among autoregressive models (\modelts, {\modeltb} and {\modeltl}), there is a perfect correlation between log-likelihood and BLEU. On all five language pairs (undistilled), the rankings of autoregressive models based on log-likelihood and BLEU are identical.
        \item Among non-autoregressive latent variable models with the same prior distribution, there is a strong but not perfect correlation. 
        Between Gauss-large and Gauss-base, the model with higher held-out log-likelihood also gives higher BLEU on four out of five datasets.
        Similarly, Flow-base gives higher log-likelihood and BLEU score than Flow-small on all datasets except {\wmtdeen}.
    \end{enumerate}
    
    \item Log-likelihood is not correlated with BLEU when comparing models from different families.
    \begin{enumerate}[
      align=left,
      leftmargin=1.0em,
      itemindent=0pt,
      labelsep=0pt,
      labelwidth=2em
    ]
    \setlength\itemsep{0.40em}
    \setlength\parskip{0.0em}
        \item Between latent variable models with different prior distributions, we observe no correlation between log-likelihood and BLEU.
        On four out of five language pairs (undistilled), Flow-base gives much higher log-likelihood but similar or worse BLEU score than Gauss-base. 
        With distillation, Gauss-large considerably outperforms Flow-base in BLEU on all datasets, while Flow-base gives better log-likelihood.
        \item Overall, autoregressive models offer the best translation quality but not the best modeling performance. In fact, Flow-base model with a non-autoregressive decoder gives the highest held-out log-likelihood on all datasets. 
    \end{enumerate}
    \vskip -0.1in
\end{enumerate}

\paragraph{Correlation between log-likelihood and BLEU across checkpoints} Table~\ref{tab:corr} presents the correlation between log-likelihood and BLEU across the training checkpoints of several models. 
The findings are similar to Table~\ref{tab:mainresult}: for Transformer-base, there is almost perfect correlation ($0.926$) across the checkpoints. 
For Gauss-base and Flow-base, we observe strong but not perfect correlation ($0.831$ and $0.678$).
Overall, these findings suggest that there is a high correlation between log-likelihood and BLEU when comparing models within the same family. We discuss the correlation for models trained with distillation below in \cref{sec:dist}.

\paragraph{Out-of-distribution experiments} We run additional experiments to validate our findings on data outside the training distribution. Using {\wmtdeen} (which is collected from news commentary and parliament proceedings) and {\iwsltdeen} (a collection of transcriptions of TED talks), we evaluate models that are trained on one dataset on the other's test set. The results, presented in Table~\ref{tab:oodresult}, are consistent with the in-distribution data.
Within the same model family (autoregressive or latent variable with the same prior distribution), the correlation between log-likelihood and BLEU is high.
Across different families of models, however, we again find no correlation. While the flow prior models are the best density estimators overall (even better than the autoregressive models), their translation quality is the poorest.
These findings show that the correlation between log-likelihood and BLEU varies significantly depending on the range of model families being compared, on both in-domain and out-of-domain data.

\subsection{Knowledge Distillation}
\label{sec:dist}
In Table~\ref{tab:corr}, we observe a strong negative correlation between log-likelihood and BLEU across the training checkpoints of several density estimators trained with distillation.
Indeed, distillation severely hurts density estimation performance on all datasets (see Tables~\ref{tab:mainresult} and \ref{tab:oodresult}).
In terms of generation quality, it consistently improves non-autoregressive models, yet the amount of improvement varies across models and datasets. 
On {\wmtende} and {\wmtdeen}, distillation gives a significant 7--9 BLEU increase for diagonal Gaussian prior models, but the improvement is relatively smaller on other datasets. Flow prior models benefit less from distillation, only 3--4 BLEU scores on {\wmtendeboth} and less on other datasets.
For autoregressive models, distillation results in a slight decrease in generation performance.

\subsection{Iterative inference on Gaussian vs. flow prior}
We analyze the effect of iterative inference on the Gaussian and the flow prior models. Table~\ref{tab:refinement} shows that iterative refinement improves BLEU and ELBO for both Gaussian prior and flow prior models, but the gain is relatively smaller for the flow prior model.

\begin{table}[!h]
\small
\centering
\vskip -0.10in
\caption{Iterative inference with a delta posterior improves BLEU and ELBO for Gauss-base and Flow-base on {\iwsltdeen} (without distillation).}
\vskip 0.15in
\begin{sc}
\begin{tabular}{llrrrrr} \toprule
& & \multicolumn{5}{c}{Number of refinement steps} \\
& & \multicolumn{1}{c}{0} & \multicolumn{1}{c}{1} & \multicolumn{1}{c}{2} & \multicolumn{1}{c}{4} & \multicolumn{1}{c}{8} \\ \midrule
\multirow{2}{*}{{\footnotesize BLEU}}
& Ga-B & 22.88 & 24.36 & 24.60 & 24.69 & 24.69 \\
& Fl-B & 24.57 & 24.89 & 24.81 & 24.92 & 24.77 \\ \midrule

\multirow{2}{*}{{\footnotesize ELBO}}
& Ga-B & -1.11 & -0.93 & -0.90 & -0.89 & -0.89 \\
& Fl-B & -1.22 & -1.17 & -1.16 & -1.15 & -1.15 \\ \bottomrule
\end{tabular}
\end{sc}
\label{tab:refinement}
\end{table}

\begin{figure}[!t]
\begin{center}
\centerline{\includegraphics[width=0.80\columnwidth]{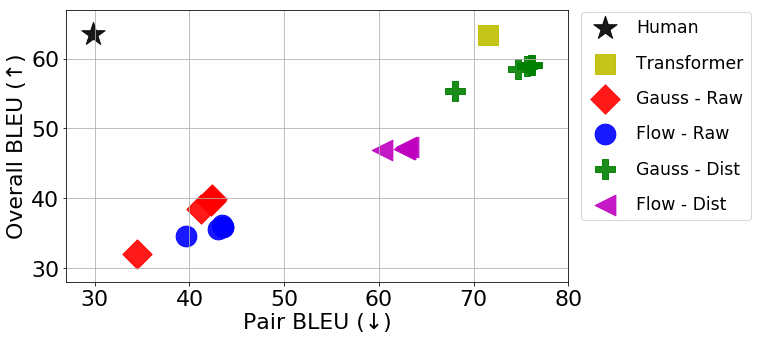}}
\caption{Quality vs. diversity analysis. we plot overall BLEU (higher means better quality overall) and pairwise-BLEU (lower means more diverse). 
For Gaussian prior and flow prior models, we perform $k$ steps of iterative inference ($k \in \{0,1,2,4,8\}$).
}
\label{fig:diversity}
\end{center}
\vskip -0.20in
\end{figure}

\paragraph{Iterative refinement improves generation quality at the cost of diversity} On the subset of {\wmtende} test set that contains 10 German references for each English sentence~\citep{ott18analyzing} we decode 10 distinct candidates from each model and compute the overall quality (BLEU) and diversity (pairwise BLEU) (see Figure~\ref{fig:diversity}). Ground-truth references, having both high quality and diversity, are in the top left. Beam candidates from an autoregressive model are of high quality but not diverse (top right). 
For the latent variable models, we compute the quality and diversity of the output after performing $k$ steps of iterative inference (where $k\in\{0,1,2,4,8\}$ and 0 indicates no refinement).
The first observation is that distillation significantly improves the quality but reduces diversity (towards top right) for non-autoregressive latent variable models. 
Iterative refinement has a similar effect for Gaussian prior models, improving quality at the cost of diversity (towards top right). For flow prior models, however, it leads to little improvement in quality and a drop in diversity (towards right).

\begin{figure}[!t]
  \centering
  \begin{minipage}[b]{0.40\textwidth}
    \includegraphics[width=\textwidth]{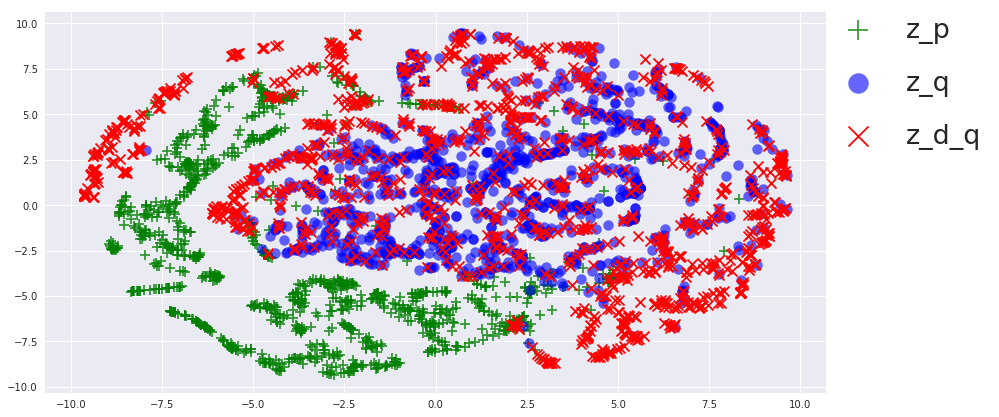}
  \end{minipage}
  \begin{minipage}[b]{0.40\textwidth}
    \includegraphics[width=\textwidth]{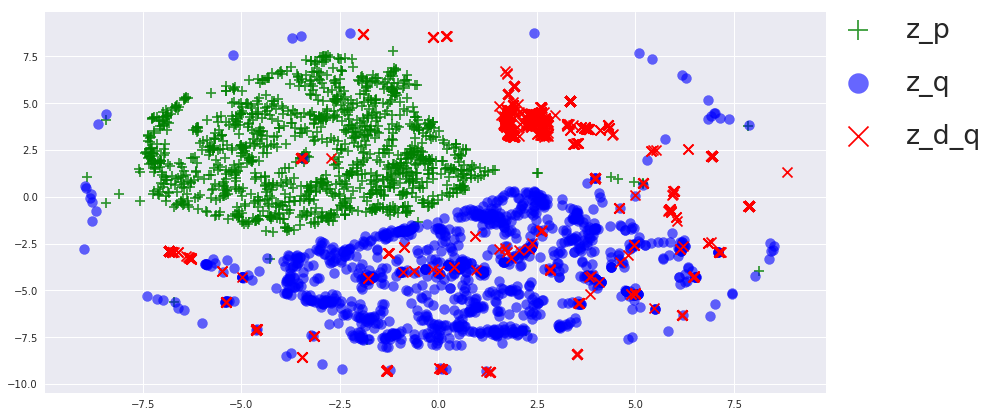}
  \end{minipage}
\caption{Visualization of the latent space with 1K samples from the prior ({\color{ForestGreen}green plus sign}), the approximate posterior ({\color{blue}blue circle}) and the delta posterior ({\color{red}red cross}) of Gauss-base (top) and Flow-small (bottom) on a {\iwsltdeen} test example.}
\label{fig:vis}
\vskip -0.1in
\end{figure}

\begin{table*}[!t]
\small
\centering
\vskip -0.10in
\caption{BLEU score, generation speed and size of various models on {\wmtende} test set. We measure generation speed in sentence/s on a single V100 GPU with batch size 1. 
We perform inference of autoregressive models using beam search with width 4.
For latent variable models, we train perform $k$ steps of iterative inference~\citep{shu19latent} (where $k\in\{0,1,2,4,8\}$) and report results from models trained with distillation.
$(*)$ results are from \citet{ma19flowseq}.}
\vskip 0.15in
\begin{sc}
\begin{tabular}{lrrrrrrrrrrrr} \toprule

& \multicolumn{5}{c}{BLEU} & & \multicolumn{5}{c}{Speed} & Size  \\
\cmidrule{2-6} \cmidrule{8-12}
$k=$ & \multicolumn{1}{c}{0} & \multicolumn{1}{c}{1} & \multicolumn{1}{c}{2} & \multicolumn{1}{c}{4} & \multicolumn{1}{c}{8} & & \multicolumn{1}{c}{0} & \multicolumn{1}{c}{1} & \multicolumn{1}{c}{2} & \multicolumn{1}{c}{4} & \multicolumn{1}{c}{8} & \\ \midrule

\modelts & 24.54 &  & & & & & 2.69 & & & & & 17M \\
\modeltb & 28.18 &  & & & & & 2.58 & & & & & 60M \\
\modeltl & 29.39 &  & & & & & 1.93 & & & & & 208M \\ \midrule
\modelgb & 23.15 & 24.54 & 24.87 & 24.94 & 24.92 & & 28.77 & 20.52 & 16.51 & 12.00 & 8.11 & 75M \\
\modelgl & 24.31 & 25.53 & 25.69 & 25.68 & 25.68 & & 19.83 & 14.72 & 10.25 & 7.88 & 4.91 & 95M \\ 
\modelfb & 21.57 & 21.82 & 21.79 & 21.81 & 21.80 & & 5.82 & 5.60 & 4.84 & 3.60 & 3.37 & 75M \\
\modelfl$^{(*)}$ & 23.72 & & & & & & & & & & & 258M \\
\bottomrule

\end{tabular}
\end{sc}
\label{tab:speed}
\end{table*}

\paragraph{Visualization of latent space}
In Figure~\ref{fig:vis}, we visualize the latent space of the approximate prior, the prior and the delta posterior of the latent variable models using t-SNE~\citep{maaten14accelerating}. 
It is clear from the figures that the delta posterior of Gauss-base has high overlap with the approximate posterior, while the overlap is relatively low for Flow-small.
We conjecture that while the loss surface of ELBO contains many local optima that we can reach via iterative refinement, not all of them share the support of the approximate posterior density (hence correspond to data). This is particularly pronounced for the flow prior model.

\subsection{Generation speed and model size}
We compare performance, generation speed and size of various models in Table~\ref{tab:speed}. 
While autoregressive models offer the best translation quality, inference is inherently sequential and slow.
Decoding from non-autoregressive latent variable models is much more efficient, and requires constant time with respect to sequence length given parallel computation. Compared to Transformer-base, Gauss-large with 1 step of iterative inference improves generation speed by 6x, at the cost of 2.6 BLEU. On {\wmtdeen}, the performance degradation is 1.9 BLEU.
Flow prior models perform much worse than the Gaussian prior models despite having more parameters and slower generation speed.

\section{Related Work}
\label{related}


For sequence generation, the gap between log-likelihood and downstream metric has long been recognized. To address this discrepancy between density estimation and approximate inference (generation), there has largely been two lines of prior work: (1) structured perceptron training for conditional random fields~\citep{lafferty01conditional,collins02discriminative,liang06end} and (2) empirical risk minimization with approximate inference~\citep{valtchev97mmie,povey02minimum,och03minimum,fu07automatic,stoyanov11empirical,hopkins11tuning,shen16minimum}. More recent work proposed to train neural sequence models directly on task-specific losses using reinforcement learning~\citep{ranzato16sequence,bahdanau17actor,jaques17sequence} or adversarial training~\citep{goyal16professor}.


Despite such a plethora of work in bridging the gap between log-likelihood and the downstream task, the exact correlation between the two has not been established well. 
Our work investigates the correlation for neural sequence models (autoregressive models and latent variable models) in machine translation.
Among autoregressive models for open-domain dialogue, a concurrent work~\citep{adiwardana20towards} found a strong correlation between perplexity and a human evaluation metric that awards sensibleness and specificity. 
This work confirms a part of our finding that log-likelihood is highly correlated with the downstream metric when we consider models within the same family.


Our work is inspired by recent work on latent variable models for non-autoregressive neural machine translation~\citep{gu18non,lee18deterministic,kaiser18fast}. 
Specifically, we compare continuous latent variable models with a diagonal Gaussian prior~\citep{shu19latent} and a normalizing flow prior~\citep{ma19flowseq}. We find that while having an expressive prior is beneficial for density estimation, a simple prior delivers better generation quality while being smaller and faster.

\section{Conclusion}
\label{sec:conclusion}
In this work, we investigate the correlation between log-likelihood and the downstream evaluation metric for machine translation.
We train several autoregressive models and latent variable models on five language pairs from three machine translation datasets ({\wmtendeboth}, {\wmtenroboth} and {\iwsltdeen}), and find that the correlation between log-likelihood and BLEU changes drastically depending on the range of model families being compared:
Among the models within the same family, log-likelihood is highly correlated with BLEU.
Between models of different families, however, we observe no correlation:
the flow prior model gives higher held-out log-likelihood but similar or worse BLEU score than the Gaussian prior model.
Furthermore, autoregressive models give the highest BLEU scores overall but the latent variable model with a flow prior gives the highest test log-likelihoods on all datasets.

In the future, we will investigate the factors behind this discrepancy. One possibility is the inherent difficulty of inference for latent variable models, which might be resolved by designing better inference algorithms. We will also explore if the discrepancy is mainly caused by the difference in the decoding distribution (autoregressive vs. factorized) or the training objective (maximum likelihood vs. ELBO).



\section*{Acknowledgements}
We thank our colleagues at the Google Translate and Brain teams, particularly Durk Kingma, Yu Zhang, Yuan Cao and Julia Kreutzer for their feedback on the draft. JL thanks Chunting Zhou, Manoj Kumar and William Chan for helpful discussions. 
KC is supported by Samsung Advanced Institute of Technology (Next Generation Deep Learning: from pattern recognition to AI) and Samsung Research (Improving Deep Learning using Latent Structure). KC also thanks eBay and NVIDIA for their support.

\bibliography{main}

\begin{thebibliography}{58}
\providecommand{\natexlab}[1]{#1}
\providecommand{\url}[1]{\texttt{#1}}
\expandafter\ifx\csname urlstyle\endcsname\relax
  \providecommand{\doi}[1]{doi: #1}\else
  \providecommand{\doi}{doi: \begingroup \urlstyle{rm}\Url}\fi

\bibitem[Adiwardana et~al.(2020)Adiwardana, Luong, So, Hall, Fiedel, Thoppilan,
  Yang, Kulshreshtha, Nemade, Lu, and Le]{adiwardana20towards}
Adiwardana, D., Luong, M.-T., So, D.~R., Hall, J., Fiedel, N., Thoppilan, R.,
  Yang, Z., Kulshreshtha, A., Nemade, G., Lu, Y., and Le, Q.~V.
\newblock Towards a human-like open-domain chatbot.
\newblock \emph{arXiv preprint arxiv:2001.09977}, 2020.

\bibitem[Bahdanau et~al.(2015)Bahdanau, Cho, and Bengio]{bahdanau15neural}
Bahdanau, D., Cho, K., and Bengio, Y.
\newblock Neural machine translation by jointly learning to align and
  translate.
\newblock In \emph{3rd International Conference on Learning Representations,
  {ICLR}}, 2015.

\bibitem[Bahdanau et~al.(2017)Bahdanau, Brakel, Xu, Goyal, Lowe, Pineau,
  Courville, and Bengio]{bahdanau17actor}
Bahdanau, D., Brakel, P., Xu, K., Goyal, A., Lowe, R., Pineau, J., Courville,
  A.~C., and Bengio, Y.
\newblock An actor-critic algorithm for sequence prediction.
\newblock In \emph{5th International Conference on Learning Representations,
  {ICLR}}, 2017.

\bibitem[Banerjee \& Lavie(2005)Banerjee and Lavie]{banerjee05meteor}
Banerjee, S. and Lavie, A.
\newblock Meteor: An automatic metric for mt evaluation with improved
  correlation with human judgments.
\newblock 01 2005.

\bibitem[Bauer \& Mnih(2019)Bauer and Mnih]{bauer19resampled}
Bauer, M. and Mnih, A.
\newblock Resampled priors for variational autoencoders.
\newblock In \emph{The 22nd International Conference on Artificial Intelligence
  and Statistics, {AISTATS}}, pp.\  66--75, 2019.

\bibitem[Bowman et~al.(2016)Bowman, Vilnis, Vinyals, Dai, J{\'{o}}zefowicz, and
  Bengio]{bowman16generating}
Bowman, S.~R., Vilnis, L., Vinyals, O., Dai, A.~M., J{\'{o}}zefowicz, R., and
  Bengio, S.
\newblock Generating sentences from a continuous space.
\newblock In \emph{Proceedings of the 20th {SIGNLL} Conference on Computational
  Natural Language Learning, CoNLL}, pp.\  10--21, 2016.

\bibitem[Chung et~al.(2014)Chung, G{\"{u}}l{\c{c}}ehre, Cho, and
  Bengio]{chung14empirical}
Chung, J., G{\"{u}}l{\c{c}}ehre, {\c{C}}., Cho, K., and Bengio, Y.
\newblock Empirical evaluation of gated recurrent neural networks on sequence
  modeling.
\newblock \emph{arXiv preprint arxiv:1412.3555}, 2014.

\bibitem[Collins(2002)]{collins02discriminative}
Collins, M.
\newblock Discriminative training methods for hidden {M}arkov models: Theory
  and experiments with perceptron algorithms.
\newblock In \emph{Proceedings of the 2002 Conference on Empirical Methods in
  Natural Language Processing ({EMNLP} 2002)}, pp.\  1--8. Association for
  Computational Linguistics, 2002.

\bibitem[Dinh et~al.(2017)Dinh, Sohl{-}Dickstein, and Bengio]{dinh17density}
Dinh, L., Sohl{-}Dickstein, J., and Bengio, S.
\newblock Density estimation using real {NVP}.
\newblock In \emph{International Conference on Learning Representations}, 2017.

\bibitem[Elman(1990)]{elman90finding}
Elman, J.~L.
\newblock Finding structure in time.
\newblock \emph{Cognitive Science}, 14\penalty0 (2):\penalty0 179--211, 1990.

\bibitem[Gehring et~al.(2017)Gehring, Auli, Grangier, Yarats, and
  Dauphin]{gehring17convolutional}
Gehring, J., Auli, M., Grangier, D., Yarats, D., and Dauphin, Y.~N.
\newblock Convolutional sequence to sequence learning.
\newblock In \emph{Proceedings of the 34th International Conference on Machine
  Learning, {ICML}}, pp.\  1243--1252, 2017.

\bibitem[Goyal et~al.(2016)Goyal, Lamb, Zhang, Zhang, Courville, and
  Bengio]{goyal16professor}
Goyal, A., Lamb, A., Zhang, Y., Zhang, S., Courville, A.~C., and Bengio, Y.
\newblock Professor forcing: {A} new algorithm for training recurrent networks.
\newblock In \emph{Advances in Neural Information Processing Systems 29: Annual
  Conference on Neural Information Processing Systems}, pp.\  4601--4609, 2016.

\bibitem[Gu et~al.(2018)Gu, Bradbury, Xiong, Li, and Socher]{gu18non}
Gu, J., Bradbury, J., Xiong, C., Li, V. O.~K., and Socher, R.
\newblock Non-autoregressive neural machine translation.
\newblock In \emph{6th International Conference on Learning Representations,
  {ICLR}}, 2018.

\bibitem[Higgins et~al.(2017)Higgins, Matthey, Pal, Burgess, Glorot, Botvinick,
  Mohamed, and Lerchner]{higgins17beta}
Higgins, I., Matthey, L., Pal, A., Burgess, C., Glorot, X., Botvinick, M.,
  Mohamed, S., and Lerchner, A.
\newblock beta-vae: Learning basic visual concepts with a constrained
  variational framework.
\newblock In \emph{5th International Conference on Learning Representations,
  {ICLR}}, 2017.

\bibitem[Hochreiter \& Schmidhuber(1997)Hochreiter and
  Schmidhuber]{hochreiter97long}
Hochreiter, S. and Schmidhuber, J.
\newblock Long short-term memory.
\newblock \emph{Neural Computation}, 9\penalty0 (8):\penalty0 1735--1780, 1997.

\bibitem[Hoffman \& Johnson(2016)Hoffman and Johnson]{hoffman16elbo}
Hoffman, M.~D. and Johnson, M.~J.
\newblock Elbo surgery: yet another way to carve up the variational evidence
  lower bound.
\newblock \emph{Workshop in Advances in Approximate Bayesian Inference,
  Neurips}, 2016.

\bibitem[Hoogeboom et~al.(2019)Hoogeboom, Peters, van~den Berg, and
  Welling]{hoogeboom19integer}
Hoogeboom, E., Peters, J. W.~T., van~den Berg, R., and Welling, M.
\newblock Integer discrete flows and lossless compression.
\newblock In \emph{Advances in Neural Information Processing Systems 32: Annual
  Conference on Neural Information Processing Systems 2019}, pp.\
  12134--12144, 2019.

\bibitem[Hopkins \& May(2011)Hopkins and May]{hopkins11tuning}
Hopkins, M. and May, J.
\newblock Tuning as ranking.
\newblock In \emph{Proceedings of the 2011 Conference on Empirical Methods in
  Natural Language Processing}, pp.\  1352--1362. Association for Computational
  Linguistics, 2011.

\bibitem[Jaques et~al.(2017)Jaques, Gu, Bahdanau, Hern{\'{a}}ndez{-}Lobato,
  Turner, and Eck]{jaques17sequence}
Jaques, N., Gu, S., Bahdanau, D., Hern{\'{a}}ndez{-}Lobato, J.~M., Turner,
  R.~E., and Eck, D.
\newblock Sequence tutor: Conservative fine-tuning of sequence generation
  models with kl-control.
\newblock In \emph{Proceedings of the 34th International Conference on Machine
  Learning, {ICML}}, pp.\  1645--1654, 2017.

\bibitem[Kaiser et~al.(2018)Kaiser, Bengio, Roy, Vaswani, Parmar, Uszkoreit,
  and Shazeer]{kaiser18fast}
Kaiser, L., Bengio, S., Roy, A., Vaswani, A., Parmar, N., Uszkoreit, J., and
  Shazeer, N.
\newblock Fast decoding in sequence models using discrete latent variables.
\newblock In \emph{Proceedings of the 35th International Conference on Machine
  Learning, {ICML}}, pp.\  2395--2404, 2018.

\bibitem[Kim \& Rush(2016)Kim and Rush]{kim16sequence}
Kim, Y. and Rush, A.~M.
\newblock Sequence-level knowledge distillation.
\newblock In \emph{Proceedings of the 2016 Conference on Empirical Methods in
  Natural Language Processing, {EMNLP}}, pp.\  1317--1327, 2016.

\bibitem[Kingma \& Ba(2015)Kingma and Ba]{kingma15adam}
Kingma, D.~P. and Ba, J.
\newblock Adam: {A} method for stochastic optimization.
\newblock In \emph{3rd International Conference on Learning Representations,
  {ICLR}}, 2015.

\bibitem[Kingma \& Dhariwal(2018)Kingma and Dhariwal]{kingma18glow}
Kingma, D.~P. and Dhariwal, P.
\newblock Glow: Generative flow with invertible 1x1 convolutions.
\newblock In \emph{Advances in Neural Information Processing Systems 31: Annual
  Conference on Neural Information Processing Systems}, pp.\  10236--10245,
  2018.

\bibitem[Kingma \& Welling(2014)Kingma and Welling]{kingma14auto}
Kingma, D.~P. and Welling, M.
\newblock Auto-encoding variational bayes.
\newblock In \emph{2nd International Conference on Learning Representations,
  {ICLR} 2014, Banff, AB, Canada, April 14-16, 2014, Conference Track
  Proceedings}, 2014.

\bibitem[Lafferty et~al.(2001)Lafferty, McCallum, and
  Pereira]{lafferty01conditional}
Lafferty, J.~D., McCallum, A., and Pereira, F. C.~N.
\newblock Conditional random fields: Probabilistic models for segmenting and
  labeling sequence data.
\newblock In \emph{Proceedings of the Eighteenth International Conference on
  Machine Learning {(ICML} 2001)}, pp.\  282--289, 2001.

\bibitem[Lee et~al.(2018)Lee, Mansimov, and Cho]{lee18deterministic}
Lee, J., Mansimov, E., and Cho, K.
\newblock Deterministic non-autoregressive neural sequence modeling by
  iterative refinement.
\newblock In \emph{Proceedings of the 2018 Conference on Empirical Methods in
  Natural Language Processing}, pp.\  1173--1182, 2018.

\bibitem[Liang et~al.(2006)Liang, Bouchard-C{\^o}t{\'e}, Klein, and
  Taskar]{liang06end}
Liang, P., Bouchard-C{\^o}t{\'e}, A., Klein, D., and Taskar, B.
\newblock An end-to-end discriminative approach to machine translation.
\newblock In \emph{Proceedings of the 21st International Conference on
  Computational Linguistics and 44th Annual Meeting of the Association for
  Computational Linguistics}, pp.\  761--768, 2006.

\bibitem[Ma et~al.(2019)Ma, Zhou, Li, Neubig, and Hovy]{ma19flowseq}
Ma, X., Zhou, C., Li, X., Neubig, G., and Hovy, E.~H.
\newblock Flowseq: Non-autoregressive conditional sequence generation with
  generative flow.
\newblock \emph{arXiv preprint arxiv:1909.02480}, 2019.

\bibitem[Mansimov et~al.(2019)Mansimov, Mahmood, Kang, and
  Cho]{mansimov19molecular}
Mansimov, E., Mahmood, O., Kang, S., and Cho, K.
\newblock Molecular geometry prediction using a deep generative graph neural
  network.
\newblock \emph{arXiv preprint arxiv:1904.00314}, 2019.

\bibitem[Och(2003)]{och03minimum}
Och, F.~J.
\newblock Minimum error rate training in statistical machine translation.
\newblock In \emph{Proceedings of the 41st Annual Meeting of the Association
  for Computational Linguistics}, pp.\  160--167, 2003.

\bibitem[Ott et~al.(2018)Ott, Auli, Grangier, and Ranzato]{ott18analyzing}
Ott, M., Auli, M., Grangier, D., and Ranzato, M.
\newblock Analyzing uncertainty in neural machine translation.
\newblock In \emph{Proceedings of the 35th International Conference on Machine
  Learning, {ICML}}, pp.\  3953--3962, 2018.

\bibitem[Papamakarios et~al.(2019)Papamakarios, Nalisnick, Rezende, Mohamed,
  and Lakshminarayanan]{papa19normalizing}
Papamakarios, G., Nalisnick, E.~T., Rezende, D.~J., Mohamed, S., and
  Lakshminarayanan, B.
\newblock Normalizing flows for probabilistic modeling and inference.
\newblock \emph{arXiv preprint arxiv:1912.02762}, 2019.

\bibitem[Papineni et~al.(2002)Papineni, Roukos, Ward, and Zhu]{papieni02bleu}
Papineni, K., Roukos, S., Ward, T., and Zhu, W.
\newblock Bleu: a method for automatic evaluation of machine translation.
\newblock In \emph{Proceedings of the 40th Annual Meeting of the Association
  for Computational Linguistics}, pp.\  311--318, 2002.

\bibitem[{Povey} \& {Woodland}(2002){Povey} and {Woodland}]{povey02minimum}
{Povey}, D. and {Woodland}, P.~C.
\newblock Minimum phone error and i-smoothing for improved discriminative
  training.
\newblock In \emph{2002 IEEE International Conference on Acoustics, Speech, and
  Signal Processing}, volume~1, pp.\  I--105--I--108, 2002.

\bibitem[{Qiang Fu} \& {Biing-Hwang Juang}(2007){Qiang Fu} and {Biing-Hwang
  Juang}]{fu07automatic}
{Qiang Fu} and {Biing-Hwang Juang}.
\newblock Automatic speech recognition based on weighted minimum classification
  error (w-mce) training method.
\newblock In \emph{2007 IEEE Workshop on Automatic Speech Recognition
  Understanding (ASRU)}, pp.\  278--283, 2007.

\bibitem[Ranzato et~al.(2016)Ranzato, Chopra, Auli, and
  Zaremba]{ranzato16sequence}
Ranzato, M., Chopra, S., Auli, M., and Zaremba, W.
\newblock Sequence level training with recurrent neural networks.
\newblock In \emph{4th International Conference on Learning Representations,
  {ICLR}}, 2016.

\bibitem[Rezende \& Mohamed(2015)Rezende and Mohamed]{rezende15variational}
Rezende, D.~J. and Mohamed, S.
\newblock Variational inference with normalizing flows.
\newblock In \emph{Proceedings of the 32nd International Conference on Machine
  Learning}, pp.\  1530--1538, 2015.

\bibitem[Rosca et~al.(2018)Rosca, Lakshminarayanan, and
  Mohamed]{rosca18distribution}
Rosca, M., Lakshminarayanan, B., and Mohamed, S.
\newblock Distribution matching in variational inference.
\newblock \emph{arXiv preprint arxiv:1802.06847}, 2018.

\bibitem[Salimans \& Kingma(2016)Salimans and Kingma]{salimans16weight}
Salimans, T. and Kingma, D.~P.
\newblock Weight normalization: {A} simple reparameterization to accelerate
  training of deep neural networks.
\newblock In \emph{Advances in Neural Information Processing Systems 29}, pp.\
  901, 2016.

\bibitem[Schuster \& Nakajima(2012)Schuster and Nakajima]{schuster12japanese}
Schuster, M. and Nakajima, K.
\newblock Japanese and korean voice search.
\newblock In \emph{2012 {IEEE} International Conference on Acoustics, Speech
  and Signal Processing, {ICASSP}}, pp.\  5149--5152, 2012.

\bibitem[Sennrich et~al.(2016)Sennrich, Haddow, and Birch]{sennrich16edinburgh}
Sennrich, R., Haddow, B., and Birch, A.
\newblock Edinburgh neural machine translation systems for {WMT} 16.
\newblock In \emph{Proceedings of the First Conference on Machine Translation,
  {WMT}}, pp.\  371--376, 2016.

\bibitem[Shen et~al.(2016)Shen, Cheng, He, He, Wu, Sun, and Liu]{shen16minimum}
Shen, S., Cheng, Y., He, Z., He, W., Wu, H., Sun, M., and Liu, Y.
\newblock Minimum risk training for neural machine translation.
\newblock In \emph{Proceedings of the 54th Annual Meeting of the Association
  for Computational Linguistics (Volume 1: Long Papers)}, pp.\  1683--1692,
  2016.

\bibitem[Shen et~al.(2019)Shen, Ott, Auli, and Ranzato]{shen19mixture}
Shen, T., Ott, M., Auli, M., and Ranzato, M.
\newblock Mixture models for diverse machine translation: Tricks of the trade.
\newblock In \emph{Proceedings of the 36th International Conference on Machine
  Learning}, pp.\  5719--5728, 2019.

\bibitem[Shu et~al.(2019)Shu, Lee, Nakayama, and Cho]{shu19latent}
Shu, R., Lee, J., Nakayama, H., and Cho, K.
\newblock Latent-variable non-autoregressive neural machine translation with
  deterministic inference using a delta posterior.
\newblock \emph{arXiv preprint arxiv:1908.07181}, 2019.

\bibitem[Stoyanov et~al.(2011)Stoyanov, Ropson, and
  Eisner]{stoyanov11empirical}
Stoyanov, V., Ropson, A., and Eisner, J.
\newblock Empirical risk minimization of graphical model parameters given
  approximate inference, decoding, and model structure.
\newblock In \emph{Proceedings of the Fourteenth International Conference on
  Artificial Intelligence and Statistics, {AISTATS}}, pp.\  725--733, 2011.

\bibitem[Sutskever et~al.(2014)Sutskever, Vinyals, and Le]{sutskever14sequence}
Sutskever, I., Vinyals, O., and Le, Q.~V.
\newblock Sequence to sequence learning with neural networks.
\newblock In \emph{Advances in Neural Information Processing Systems 27: Annual
  Conference on Neural Information Processing Systems}, pp.\  3104--3112, 2014.

\bibitem[Tabak \& Turner(2013)Tabak and Turner]{tabak13family}
Tabak, E.~G. and Turner, C.~V.
\newblock A family of nonparametric density estimation algorithms.
\newblock \emph{Communications on Pure and Applied Mathematics}, 66\penalty0
  (2):\penalty0 145--164, 2013.

\bibitem[Tomczak \& Welling(2018)Tomczak and Welling]{tomczak18vae}
Tomczak, J.~M. and Welling, M.
\newblock {VAE} with a vampprior.
\newblock In \emph{International Conference on Artificial Intelligence and
  Statistics, {AISTATS}}, pp.\  1214--1223, 2018.

\bibitem[Tran et~al.(2019)Tran, Vafa, Agrawal, Dinh, and Poole]{tran19discrete}
Tran, D., Vafa, K., Agrawal, K.~K., Dinh, L., and Poole, B.
\newblock Discrete flows: Invertible generative models of discrete data.
\newblock In \emph{Advances in Neural Information Processing Systems 32: Annual
  Conference on Neural Information Processing Systems 2019}, pp.\
  14692--14701, 2019.

\bibitem[Valtchev et~al.(1997)Valtchev, Odell, Woodland, and
  Young]{valtchev97mmie}
Valtchev, V., Odell, J.~J., Woodland, P.~C., and Young, S.~J.
\newblock Mmie training of large vocabulary recognition systems.
\newblock \emph{Speech Commun.}, 22\penalty0 (4):\penalty0 303–314, 1997.

\bibitem[van~den Oord et~al.(2016)van~den Oord, Dieleman, Zen, Simonyan,
  Vinyals, Graves, Kalchbrenner, Senior, and Kavukcuoglu]{oord16wavenet}
van~den Oord, A., Dieleman, S., Zen, H., Simonyan, K., Vinyals, O., Graves, A.,
  Kalchbrenner, N., Senior, A.~W., and Kavukcuoglu, K.
\newblock Wavenet: {A} generative model for raw audio.
\newblock In \emph{The 9th {ISCA} Speech Synthesis Workshop}, pp.\  125, 2016.

\bibitem[van~den Oord et~al.(2018)van~den Oord, Li, Babuschkin, Simonyan,
  Vinyals, Kavukcuoglu, van~den Driessche, Lockhart, Cobo, Stimberg,
  Casagrande, Grewe, Noury, Dieleman, Elsen, Kalchbrenner, Zen, Graves, King,
  Walters, Belov, and Hassabis]{oord18parallel}
van~den Oord, A., Li, Y., Babuschkin, I., Simonyan, K., Vinyals, O.,
  Kavukcuoglu, K., van~den Driessche, G., Lockhart, E., Cobo, L.~C., Stimberg,
  F., Casagrande, N., Grewe, D., Noury, S., Dieleman, S., Elsen, E.,
  Kalchbrenner, N., Zen, H., Graves, A., King, H., Walters, T., Belov, D., and
  Hassabis, D.
\newblock Parallel wavenet: Fast high-fidelity speech synthesis.
\newblock In \emph{Proceedings of the 35th International Conference on Machine
  Learning, {ICML}}, pp.\  3915--3923, 2018.

\bibitem[van~der Maaten(2014)]{maaten14accelerating}
van~der Maaten, L.
\newblock Accelerating t-sne using tree-based algorithms.
\newblock \emph{J. Mach. Learn. Res.}, 15\penalty0 (1):\penalty0 3221--3245,
  2014.

\bibitem[Vaswani et~al.(2017)Vaswani, Shazeer, Parmar, Uszkoreit, Jones, Gomez,
  Kaiser, and Polosukhin]{vaswani17attention}
Vaswani, A., Shazeer, N., Parmar, N., Uszkoreit, J., Jones, L., Gomez, A.~N.,
  Kaiser, L., and Polosukhin, I.
\newblock Attention is all you need.
\newblock In \emph{Advances in Neural Information Processing Systems 30: Annual
  Conference on Neural Information Processing Systems}, pp.\  5998--6008, 2017.

\bibitem[Vinyals \& Le(2015)Vinyals and Le]{vinyals15neural}
Vinyals, O. and Le, Q.~V.
\newblock A neural conversational model.
\newblock \emph{arXiv preprint arxiv:1506.05869}, 2015.

\bibitem[Vinyals et~al.(2015)Vinyals, Toshev, Bengio, and Erhan]{vinyals15show}
Vinyals, O., Toshev, A., Bengio, S., and Erhan, D.
\newblock Show and tell: {A} neural image caption generator.
\newblock In \emph{{IEEE} Conference on Computer Vision and Pattern
  Recognition, {CVPR}}, pp.\  3156--3164, 2015.

\bibitem[Wainwright \& Jordan(2008)Wainwright and
  Jordan]{wainwright08graphical}
Wainwright, M.~J. and Jordan, M.~I.
\newblock Graphical models, exponential families, and variational inference.
\newblock \emph{Foundations and Trends in Machine Learning}, 1\penalty0
  (1-2):\penalty0 1--305, 2008.

\bibitem[Zhou et~al.(2019)Zhou, Neubig, and Gu]{zhou19understanding}
Zhou, C., Neubig, G., and Gu, J.
\newblock Understanding knowledge distillation in non-autoregressive machine
  translation.
\newblock \emph{arXiv preprint arxiv:1911.02727}, 2019.

\end{thebibliography}
\bibliographystyle{icml2020}


\end{document}